\documentclass[lettersize,journal]{IEEEtran}
\usepackage{amsmath,amsfonts}
\usepackage{algorithmic}
\usepackage{algorithm}
\usepackage{booktabs} 
\usepackage{array}
\usepackage[caption=false,font=normalsize,labelfont=rm,textfont=rm]{subfig}
\usepackage{textcomp}
\usepackage{stfloats}
\usepackage{float}
\usepackage[ colorlinks = true,
linkcolor = blue,
urlcolor  = blue,
citecolor = red,
anchorcolor = green,]{hyperref}

\usepackage{verbatim}
\usepackage{graphicx}
\usepackage{epstopdf}
\usepackage{cite}
\usepackage{adjustbox}
\usepackage{float}
\usepackage{multirow}
\usepackage{pifont}
\usepackage{amssymb}

\usepackage[table,xcdraw]{xcolor}
\begin{document}

\title{
    Popeye: A Unified Visual-Language Model \\for Multi-Source Ship Detection from Remote Sensing Imagery
    \\ 
}
\author{
    Wei Zhang, Miaoxin Cai,~\IEEEmembership{Student Member,~IEEE,} Tong Zhang,~\IEEEmembership{Student Member,~IEEE,} Guoqiang Lei,\\Yin Zhuang\textsuperscript{\dag}~\IEEEmembership{Member,~IEEE,}  and Xuerui Mao\textsuperscript{\dag}  

    \thanks{ {\dag} Co-corresponding author: Yin Zhuang and Xuerui Mao.}
    \thanks{Wei Zhang and Guoqiang Lei are with the Advanced Research Institute of Multidisciplinary Sciences, Beijing Institute of Technology, Beijing 100081, China, and also with the School of Mechatronical Engineering, Beijing Institute of Technology, Beijing 100081, China. (e-mail: w.w.zhanger@gmail.com, 3120235339@bit.edu.cn).} 
    \thanks{Xuerui Mao is with the Advanced Research Institute of Multidisciplinary Sciences, Beijing Institute of Technology, Beijing 100081, China, and with the School of Mechatronical Engineering, Beijing Institute of Technology, Beijing 100081, China, and also with Yangtze Delta Region Academy of Beijing Institute of Technology, Jiaxing 314003, China. (e-mail: xmao@bit.edu.cn).
    }
    \thanks{Yin Zhuang, Miaoxin Cai, and Tong Zhang are with the National Key Laboratory of Science and Technology on Space-Born Intelligent Information Processing, Beijing Institute of Technology, Beijing 100081, China. (e-mail: yzhuang@bit.edu.cn, 3120220667@bit.edu.cn, bit\_zhangtong@163.com).
    }

}
\maketitle

\begin{abstract}
Ship detection needs to identify ship locations from remote sensing (RS) scenes. Due to different imaging payloads, various appearances of ships, and complicated background interference from the bird's eye view, it is difficult to set up a unified paradigm for achieving multi-source ship detection. To address this challenge, in this article, leveraging the large language models (LLMs)'s powerful generalization ability, a unified visual-language model called Popeye\footnote{Inspired by the animated series \textit{Popeye the Sailor}, mirroring the model's agility in tackling ship tasks within the maritime domain.} is proposed for multi-source ship detection from RS imagery. Specifically, to bridge the interpretation gap between the multi-source images for ship detection, a novel unified labeling paradigm is designed to integrate different visual modalities and the various ship detection ways, i.e., horizontal bounding box (HBB) and oriented bounding box (OBB). Subsequently, the hybrid experts encoder is designed to refine multi-scale visual features, thereby enhancing visual perception. Then, a visual-language alignment method is developed for Popeye to enhance interactive comprehension ability between visual and language content. Furthermore, an instruction adaption mechanism is proposed for transferring the pre-trained visual-language knowledge from the nature scene into the RS domain for multi-source ship detection. In addition, the segment anything model (SAM) is also seamlessly integrated into the proposed Popeye to achieve pixel-level ship segmentation without additional training costs. Finally, extensive experiments are conducted on the newly constructed ship instruction dataset named MMShip, and the results indicate that the proposed Popeye outperforms current specialist, open-vocabulary, and other visual-language models for zero-shot multi-source ship detection.
\end{abstract}


\section{Introduction}
\IEEEPARstart{S}{hip} detection in the present work refers to the technique to accurately identify ship locations from the remote sensing (RS) imagery with complex background interference \cite{han2021shipyolo}. Intelligent ship detection is critical for the analyses and monitoring of marine environments\cite{cui2019dense}. It is used in a wide range of fields such as maritime safety, border and territorial defense\cite{10547418}, naval warfare\cite{zhu2010novel}, environmental protection, search and rescue operations, maritime traffic control\cite{wang2021review}, and fishery management\cite{dong2018ship}. 

In the field of ship detection, many deep learning algorithms have been proposed. For the optical ship RS images analyses, the development of horizontal bounding box (HBB) detection methods, including CFF-SDN\cite{zhang2020intelligent} and Li \emph{et al.}\cite{li2021domain}, alongside oriented bounding box (OBB) detection methods such as RR-CNN\cite{liu2017rotated} and  Li \emph{et al.}\cite{li2020novel}, has significantly improved the accuracy of optical ship detection. For synthetic aperture radar (SAR) ship image interpretation, many HBB algorithms including CP-FCOS\cite{sun2021anchor}, SSDv2\cite{chen2020sar}, Zhu \emph{et al.}\cite{zhu2020rapid}, ARPN\cite{zhao2020attention}, and Chang \emph{et al.}\cite{chang2019ship}, as well as OBB ones such as He \emph{et al.}\cite{he2021learning} and R-FCOS\cite{zhao2021anchor} have been proposed to enhance the performance of SAR ship detection. However, those are limited to handling individual detection tasks or single-source scenarios, rendering them incapable of uniformly interpreting multi-source ship images and performing multi-granularity detection within one framework. The aforementioned characteristics show that those algorithms lack generalization ability, constraining their applications to real-world problems. 

\begin{figure}[!t]
\centering
\includegraphics[width=3.5in]{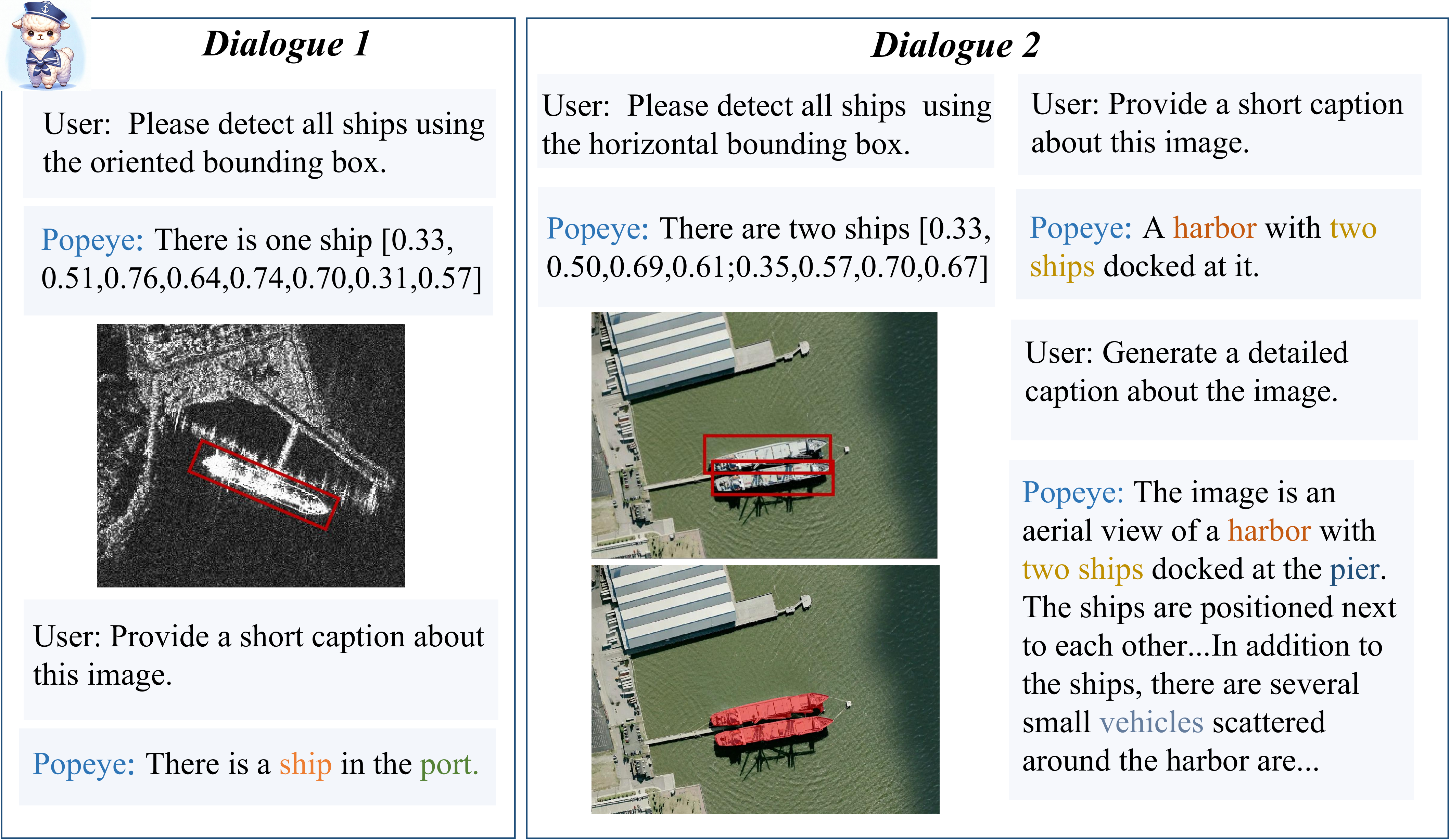}
\caption{Examples of multi-source (optical/SAR) ship image interpretation by the proposed Popeye in the multi-turn dialogue, including ship detection via OBB or HBB, segmentation, as well as
image captioning.}
\label{example1}
\end{figure}

Most recently, Large Language Models (LLMs)\cite{brown2020language,touvron2023llama,zhang2022opt} have emerged as popular and innovative tools for human assistance, exhibiting robust generalization capabilities. Among LLMs, a model named ChatGPT~\cite{openai2023chatgpt} stands out as a remarkable example, providing great potential for supporting humans in a diverse range of tasks. In addition, the considerable achievements of LLMs have sparked extensive research in introducing additional visual input and developing powerful visual-language models (VLMs)~\cite{openai2023gpt4,radford2019language}. Typically, studies such as MiniGPT-4~\cite{zhu2023minigpt}, LLaVA~\cite{liu2023visual}, and LLaMA-Adapter~\cite{zhang2023llama,gao2023llama} series have sparked a new wave of research on endowing LLMs with visual reasoning ability. The powerful capabilities of VLMs have been demonstrated in various natural scenes. However, different from natural scene images, RS ones are gathered from an overhead view by satellites. Effectively adapting current VLMs to the field of RS, and establishing a visual-language alignment paradigm for interpreting multi-source ship images present significant challenges. To address these issues, this work aims to construct a unified visual-language framework to understand multi-source and multi-modal ship data in the RS domain.

In this paper, to exploit the powerful generalization ability of LLMs for developing a universal ship detection paradigm, a novel unified VLM model named Popeye is proposed for multi-source ship interpretation in the RS domain. The proposed Popeye unifies various multi-granularity ship detection tasks, i.e., horizontal bounding box (HBB) and oriented bounding box (OBB), and integrates multi-source imagery including optical and SAR in a naive visual-language alignment procedure. As shown in Fig. \ref{example1}, Popeye can complete ship HBB/OBB detection, segmentation, and captioning tasks in multi-turn dialogues. 

Notably, based on the characteristics of RS imagery which features cluttered background and complex semantics, a hybrid experts encoder is developed. The hybrid experts encoder is utilized to refine multi-scale visual representation from input images which are pre-processed into multi-resolution, enhancing the visual perception. After that, the visual features undergo dimension transformation via the alignment projection layer, preparing for subsequent concatenation with language features to form the multi-modal input. Furthermore, to tackle the interpretation discrepancies across images from different RS visual modalities, a unified labeling paradigm is developed. Employing the proposed unified labeling method, a dataset named MMShip featuring multi-modal and multi-source instruction-following is constructed based on existing ship detection datasets.

To equip Popeye with visual and language interactive comprehension and instruction-following capability, cross-domain joint training is proposed. The first training stage leverages large-scale nature scene domain data to achieve visual-language alignment and mutual understanding. Specifically, we begin with a frozen LLaMA\cite{touvron2023llama} model, and only insert the several Low-Rank Adaptation (LoRA)~\cite{hu2021lora} metrics into LLaMA for training. Through the aforementioned process, the language-only LLM is efficiently converted into a VLM. Following that, the instruction adaption mechanism~\cite{zhang2022consecutive} is proposed to tune on the newly constructed MMShip dataset. Particularly, to further stimulate the instruction-following ability of LLM, more learnable parameters are injected into the transformer layers for training, enabling the developed VLM to adapt multi-source ship instruction detection tasks. Through step-wise and cross-domain joint training, Popeye successfully generalizes the natural domain knowledge to the ship RS domain and achieves the multi-source detection ability. In addition, Popeye is integrated with the segment anything model (SAM)~\cite{kirillov2023segment} to extend language-guided ship segmentation capability without additional training costs. The proposed model can generate the accurate bounding boxes as the SAM prompt. In conclusion, the proposed VLM Popeye can effectively unify ship HBB detection, OBB detection, and segmentation, thereby meeting the requirements of multi-granularity detection. 

Extensive experiments are conducted, demonstrating that Popeye has superior zero-shot performance on ship interpretation tasks compared to current specialist, open-vocabulary, and other visual-language models. For example, Popeye exhibits notable HBB detection improvements of 40.55\%, 26.18\%, and 4.16\% on AP@40, AP@50, and AP@60 on the DSSDD dataset compared with the best results in other Visual-language Models and open-vocabulary models. For OBB detection, it is observed Popeye achieves improvements of 4.63\% and 2.86\% on AP@40 and AP@50 with Oriented R-CNN trained on DOTA.  In conclusion, the proposed Popeye significantly lowers the resource-intensive requirement of retraining on new data, demonstrating exceptional open-domain reasoning capabilities in the ship domain.

In summary, the contributions of this paper are as follows.
    \begin{enumerate}
         \item To the best of our knowledge, a unified visual-language model named Popeye, tailored for ship imagery in the RS domain, is proposed for the first time. The proposed Popeye excels in multi-source imagery interpretation and multi-granularity ship detection tasks within the multi-turn dialogue.
        \item A cross-domain joint training strategy is designed, leveraging both nature domain caption data and RS domain instruction data to perform stage-wise parameter optimization in a lightweight manner. In the visual-language alignment stage, the LoRA tuning is adopted to transform the language-only LLM into a VLM efficiently. Subsequently, in the ship domain adaption stage, the instruction adaptor is developed to endow Popeye with task-specific instruction-following ability.
        \item A unified labeling paradigm is proposed to convert various annotation methods into the uniform image-instruction-answer data format. Leveraging the designed universal labeling paradigm, a multi-source ship instruction dataset called MMShip is constructed for the first time. MMShip contains 81k instruction data, covers HBB/OBB detection ways, and includes optical and SAR visual modalities. The construction of MMShip mitigates the challenge of lacking ship instruction datasets and facilitating the development of VLMs in the maritime domain.
        \item Extensive experiments demonstrate that Popeye excels in zero-shot ship HBB and OBB detection tasks, exceeding existing specialist, open-vocabulary, and other visual-language models. Moreover, Popeye shows excellent ship segmentation performance in challenging scenarios. Therefore, Popeye contributes a novel and generalizable visual-language paradigm for diverse multi-source ship RS imagery interpretation tasks.
\end{enumerate}

\section{Related Work}
\subsection{Large Language Models (LLMs)}
In recent years, Natural Language Processing (NLP) has made significant progress, particularly with the advent of LLMs based on Transformer architectures. Among the LLMs, The GPT series~\cite{radford2018improving,radford2019language,radford2021learning} has gained considerable attention as
a promising AI technique for NLP tasks. Especially GPT-3~\cite{brown2020language}, has demonstrated the power of massive model scaling, with models containing billions to trillions of parameters. InstructGPT~\cite{ouyang2022training} and ChatGPT~\cite{openai2023chatgpt} have shown remarkable fluency and adaptability in various conversational tasks, which has improved their ability to follow instructions. Furthermore, the open-source community has contributed resources such as LLaMA~\cite{touvron2023llama} and LLaMA-2~\cite{touvron2023llama}, enriching the LLM's instruction-following capability. Recent developments like Alpaca~\cite{taori2023alpaca}, Vicuna~\cite{chiang2023vicuna}, and GPT-4-LLM~\cite{peng2023instruction} have proposed full fine-tuning to acquire the instruction-following ability of LLMs successfully. In contrast, LoRA~\cite{hu2021lora}and LLaMA-Adapter~\cite{zhang2023llama} validate that parameter-efficient fine-tuning (PEFT) approaches can potentially replace full parameter updates during the supervised fine-tuning of LLMs. In this paper, Popeye is based on LLaMA's language understanding and inspired by superior PEFT technique to fine-tune LLMs to achieve instruction-following ability with multi-modal input.

\subsection{Visual-Language Models (VLMs)}
The fusion of LLMs and visual information revolutionizes image processing and unlocks new practical applications in various fields. Previous efforts like the VisualGPT~\cite{chen2022visualgpt} and BLIP~\cite{li2022blip} series have demonstrated the possibilities of integrating LLMs with visual inputs, showcasing their effectiveness in tasks such as image captioning and visual question answering. Recently, GPT-4~\cite{openai2023gpt4} has showcased remarkable visual instruction-following abilities by handling visual-language inputs for multi-tasks. Moreover, Bard~\cite{Google2023Bard} also has demonstrated exceptional proficiency in multi-modal understanding and reasoning across diverse tasks. Concurrently, numerous works have focused on integrating LLaMA with the vision modality to enhance visual instruction-following capabilities. Models like LLaVA~\cite{liu2023visual} and MiniGPT-4~\cite{zhu2023minigpt} assemble high-quality multi-modal instruction-following data using ChatGPT or GPT-4. They employ a simple projection layer to integrate vision encoders with LLM and fine-tune the models on the curated data. The LLaMA-Adapter V2~\cite{gao2023llama} introduces zero-initialized attention mechanisms for efficient visual instruction tuning, while mPLUG-Owl~\cite{ye2023mplug} utilizes specially designed intermediate networks for effective cross-modal alignment. The mentioned above model primarily focus on captioning and visual-question-answering (VQA) tasks.

Notably, current nature scene VLMs can also deal with object detection and segmentation tasks. For example, Lenna~\cite{wei2023lenna} develops a language-enhanced reasoning detection assistant. Sphinx~\cite{lin2023sphinx} and Qwen-VL-Chat~\cite{bai2023qwen} integrate a wide range of visual instruction tasks to tune the pre-trained model to achieve universal ability including detection. Moreover, LISA~\cite{lai2023lisa} efficiently injects segmentation capabilities into VLM, enabling reasoning segmentation guided by implicit human instructions.

Fortunately, RS can directly benefit from the existing VLMs, namely, fine-tuning VLMs to enable processing and analyzing RS images. However, the application of generalized VLMs in the RS field has been relatively limited. A notable attempt in this direction is RSGPT~\cite{hu2023rsgpt}, which aimed to develop a model capable of tackling a variety of tasks. However, each specific task requires individual fine-tuning, which makes RSGPT a poor generalization performance. In the latest update, Geochat ~\cite{kuckreja2023geochat} has introduced a more integrated approach, intending to broaden the VLM's capabilities to encompass various visual reasoning tasks including object detection. However, there is no VLM tailored for the ship domain, to unify the multi-granularity and multi-source ship detection including HBB, OBB detection, and segmentation. Therefore, this paper aims to build a unified learning frame to fill this gap.  

\subsection{Deep Learning Based Ship Object Detection}
Numerous algorithms based on deep learning have been proposed in the field of ship detection. High-performance RS object detectors often rely on the RCNN~\cite{girshick2014rich,xie2021oriented,girshick2015fast} framework, consisting of a region proposal network and regional CNN detection heads. For the universal RS object detection, Variations like the RoI transformer~\cite{ding2019learning} have been proposed, which leverages fully connected layers to rotate candidate horizontal anchor boxes before extracting features for regression and classification. Furthermore, AO2-DETR~\cite{dai2022ao2} introduces a transformer-based detection framework, which brings more research diversity. 

For the ship target detection from optical imagery, RR-CNN\cite{liu2017rotated} is a rotated region based CNN method that can accurately extract features from rotated regions and precisely locate rotated objects. CFF-SDN\cite{zhang2020intelligent} ship detection network uses multi-layer convolutional feature fusion to improve HBB high-precision ship detection. For the ship detection from SAR imagery, CP-FCOS\cite{sun2021anchor} is an anchor-free method proposed for high-resolution SAR ship images. DAPN\cite{sun2021anchor}. YOLOv2-reduced\cite{chang2019ship} architecture proposes an enhanced GPU-based deep learning method.  In addition to image-level object detection, visual grounding is a region-level task, such as CLIP-VG~\cite{xiao2023clip} and RSVG~\cite{zhan2023rsvg} can locate the referred objects described by the language instruction, attracted much attention recently.  However, Those algorithms are unable to uniformly understand multi-source ship images and complete both HBB and OBB ship detection tasks in one framework, which would constrain current intelligent interpretation methods for real-world applications. Therefore, this paper focuses on designing a unified multi-source ship RS image understanding framework, with a superior generalization capacity for cross-modal and multi-source image learning.

\begin{figure*}[!t]
\centering
\includegraphics[width=6.5in]{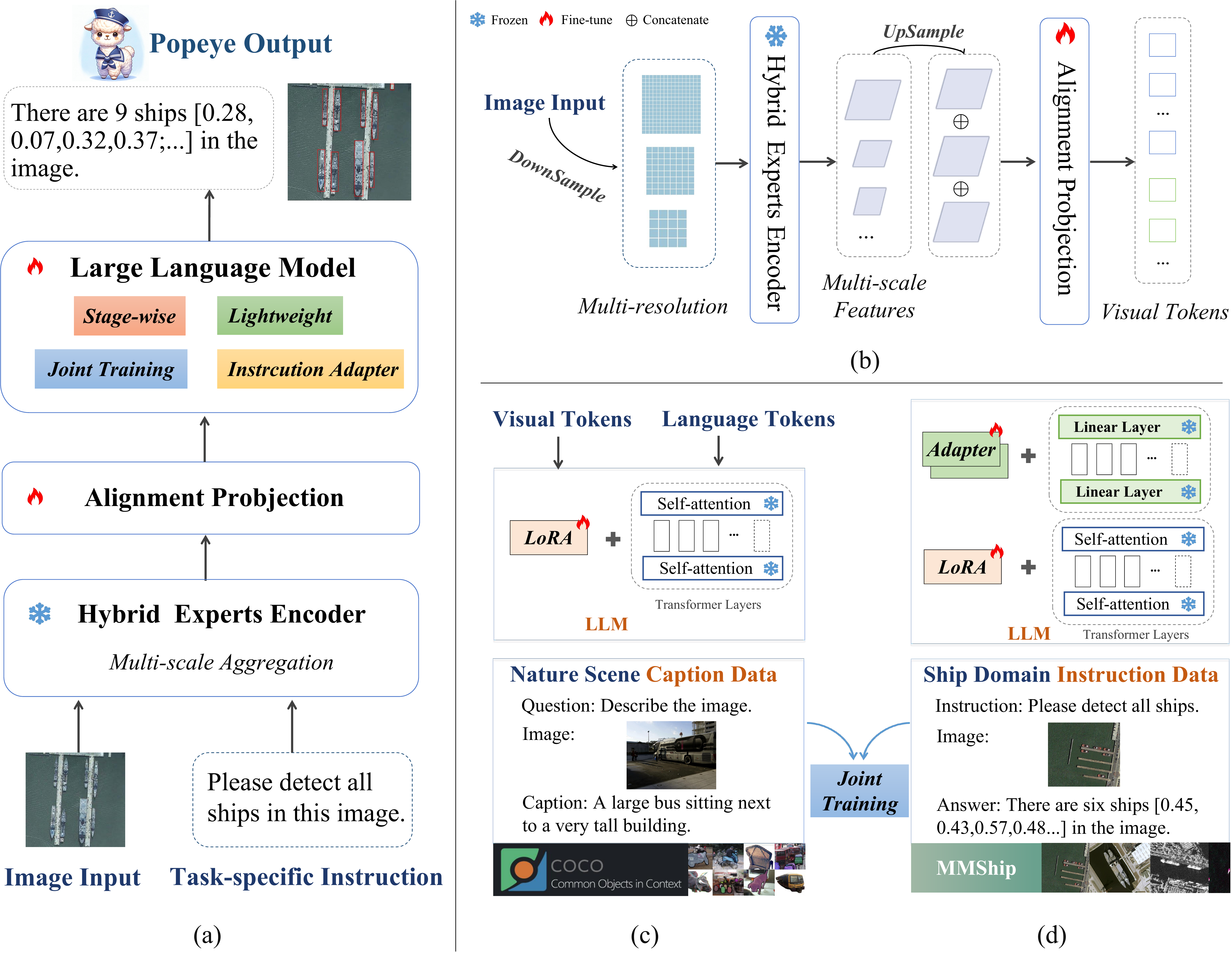}
\caption{(a) Overview of the proposed Popeye. (b) Enhanced visual perception: refining robust multi-scale visual features. (c) Visual-language alignment stage: realizing fundamental visual understanding and image-text mutual interaction. (d) Instruction adaption mechanism: achieving instruction-following ability in the ship domain.}
\label{overall}
\end{figure*}

\section{Methods}
The entire model of Popeye is summarized in Section III-A. Subsequently, the enhanced visual perception module is elaborated in Section III-B. Then, the visual-language alignment method and the instruction adaption mechanism are presented respectively in Sections III-C and III-D. Finally, we detail the integration with SAM in Section III-E. 

\subsection{Overview}

Constructing a universal model tailored to the multi-source ship imagery in the RS domain is indispensable. To this end, a unified visual-language model called Popeye is proposed for multi-granularity ship detection from different visual imagery modalities, the overview is illustrated in Fig. \ref{overall} (a). Since RS imagery is scale-variation and features cluttered background that leads to hard comprehension, a novel hybrid experts encoder is designed to refine robust visual features. Moreover, to capture the rich contextual dependencies between semantically salient regions, the input images are pre-processed into multi-resolutions and fed to the hybrid experts encoder to refine multi-scale visual features, which enables our model to achieve competitive performance. Subsequently, by employing the alignment projection layer, the multi-scale visual features are transformed into one-dimensional vectors, preparing for the subsequent integration with language vectors to form multi-modal input.

In order to endow the LLM with fundamental image understanding and instruction-following capability, a cross-domain joint training strategy for Popeye is proposed. Specifically, both nature scene image-text caption data and RS ship instruction data are leveraged for training. The cross-domain joint training can enable Popeye to have robust zero-shot reasoning ability. Furthermore, to efficiently transform language-only LLMs into a multi-modal instruction-following model, an instruction adapter is designed and the LoRA technique is employed to fine-tune LLaMA by stages, instead of updating the full parameters. Based on the aforementioned designed methods, when inputting images and task-specific instructions, Popeye can generate textual responses to accomplish ship-related tasks. The details of the network are elaborated as follows.

\subsection{Enhanced Visual Perception}

Given that RS images captured from an overhead perspective inherently including complex background interference leads to significant challenges in accurate processing. To address this challenge and better utilize visual scale information, an enhanced visual perception module is proposed. As depicted in Fig. \ref{overall} (b), the input images are downsampled to multi-resolutions, and fed to the hybrid experts encoder. The hybrid experts encoder contains various frozen visual backbones for image encoding, providing complementary visual semantics. After image encoding, the multi-scale features are extracted, then they are upsampled to the same dimensions for subsequent concatenation. The next step is multi-scale feature aggregation to obtain global robust visual representations. Particularly, The CLIP ViT-L/14 \cite{radford2021learning} backbone is adopted to extract multi-scale visual features from each input image $I$. Token embeddings generated by CLIP encoder denoted as $ \left \{ {f_{v}^{i} } \right \} _{i=1}^n \in \mathbb{R} ^{H \times W \times C}  $, where $H\times W \times C$ is the input image resolution, $n$ is the scale number. Then the multi-scale feature is denoted as $F_v$. The entire process can be formulated as
\begin{equation}
\label{deqn_ex1a}
{f_{v}^{i}} = \mathrm{CLIP_{enc}} (I_i),
\end{equation}
\begin{equation}
\label{deqn_ex1a}
F_{v} = \mathrm{Concat}\left [ f_{v}^{1},f_{v}^{2},...,f_{v}^{n} \right ]. 
\end{equation}
The DINOv2 ViT-L/14\cite{oquab2023dinov2} is further employed as another vision backbone to learn multi-scale visual tokens. Token embeddings generated by DINOv2 encoder is denoted as $ \left \{ {g_{v}^{i} } \right \} _{i=1}^m \in \mathbb{R} ^{H\times W\times C}  $, where $m$ represents the scale number. Then the multi-scale feature is denoted as $G_v$. Similarly to the previous step, formulated as
\begin{equation}
\label{deqn_ex1a}
{g_{v}^{i}} = \mathrm{DINOv2_{enc}} (I_i),
\end{equation}
\begin{equation}
\label{deqn_ex1a}
G_{v} = \mathrm{Concat}\left [ g_{v}^{1},g_{v}^{2},...,g_{v}^{m} \right ]. 
\end{equation}
Then, the features extracted in the previous two steps are concatenated along the same channel dimension. Following that, the alignment projection layer is used for dimension alignment with language tokens, namely, the aggregated visual features are projected into one-dimensional visual tokens. The visual tokens are represented as $p_v\in \mathbb{R}^{1\times C}$. The process can be expressed as 
\begin{equation}
\label{deqn_ex1a}
{p_v} =\mathrm{Projection} (\mathrm{Concat} \left [F_{v},G_{v}  \right ] ).
\end{equation}
During the enhanced visual perception process, the vision backbones in the hybrid experts encoder keep frozen to refine coarse-scale semantic information and fine-scale detail visual information.

\begin{figure*}[!t]
\centering
\includegraphics[width=6.5in]{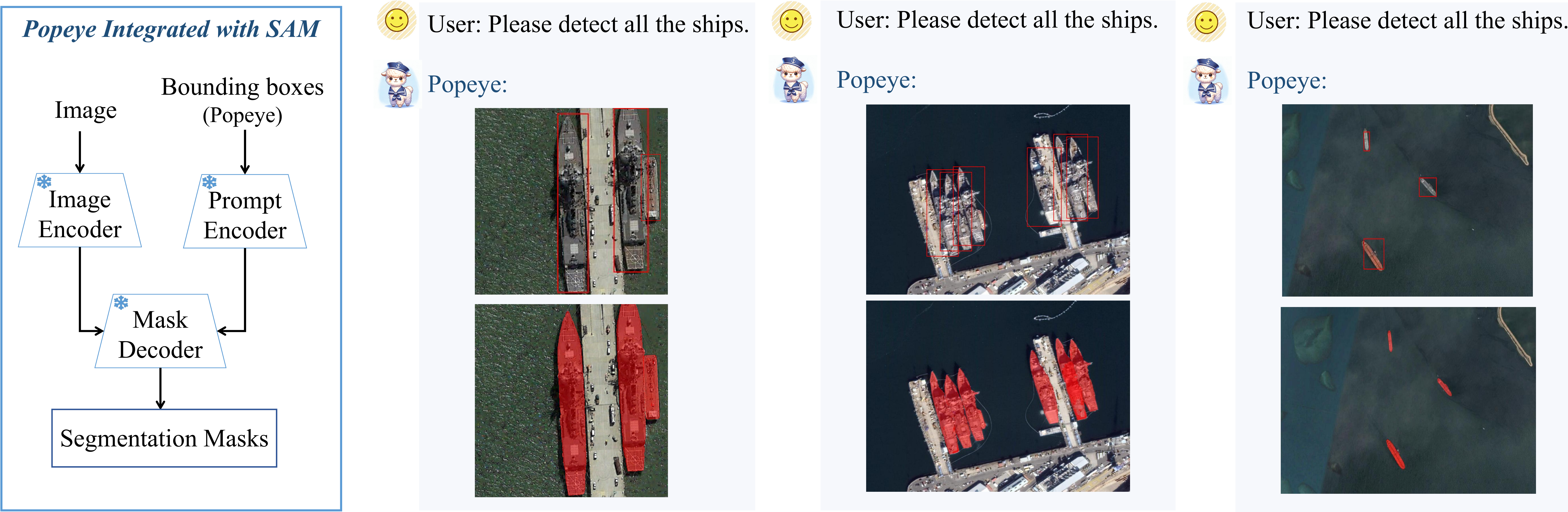}
\caption{Integrated with SAM and the examples of language-referred pixel-level segmentation.}
\label{example}
\end{figure*}

\begin{figure*}[!htbp]
\centering
\includegraphics[width=6.5in]{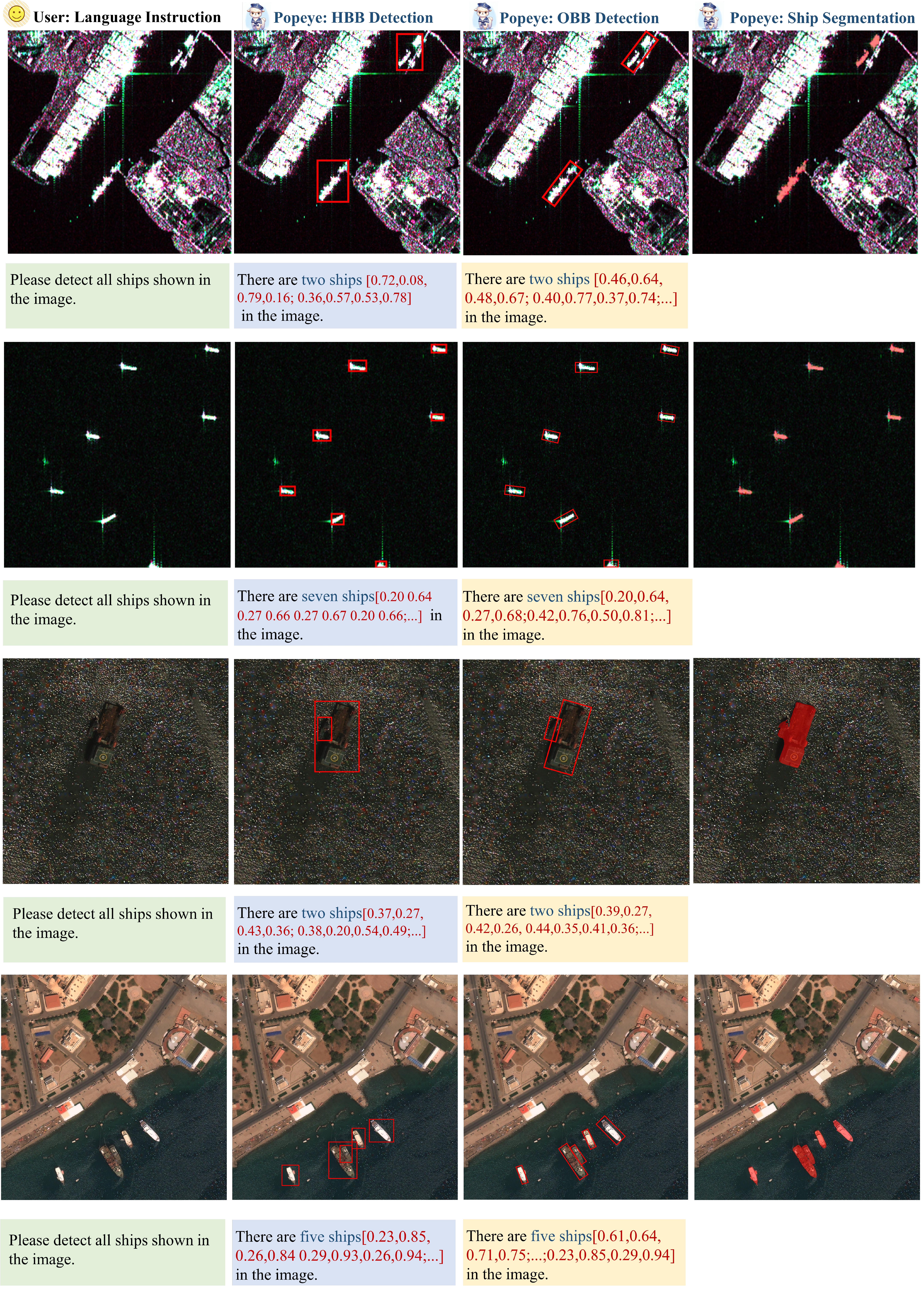}
\caption{Examples of Popeye for ships interpretation from more challenging SAR and optical RS imagery in ShipRSimagenet and DSSDD datasets. From left to right displays the results of Popeye for OBB detection, HBB detection, and ship instance segmentation of small and blurred ship targets. }
\label{example2}
\end{figure*}

\subsection{Visual-Language Alignment}

The aim of this part is to realize visual-language alignment to leverage language for facilitating image understanding. The visual features are refined by the aforementioned step, simultaneously, following most natural language processing (NLP) models\cite{touvron2023llama}, the language tokenizer is employed to embed the language instructions into discrete tokens. The language tokens are denoted as $p_l\in \mathbb{R}^{1\times C }$. By directly concatenating $p_v$ and $p_l$, a multi-modal input ${\mathcal X}$ is obtained. This process can be expressed as
\begin{equation}
\label{def_x}
{\mathcal X} =\mathrm{Concat} [\overbrace {p_{v}^{1},p_{v}^{2},~...~,p_{v}^{N_v}}^{{\rm{visual~tokens~\mathit{p_v} }}},\underbrace {p_{l}^{1},p_{l}^{2},~...~,p_{l}^{N_l}}_{\mathrm{language~tokens} ~{p_l}}],
\end{equation}
where $N_v$ represents the token length of visual features, $N_l$ denotes the token length of language features, $(p_{v}^{1},p_{v}^{2}, ~...~,p_{v}^{N_v})$ are the mixed visual backbone tokens from $p_v$, and $(p_{l}^{1},p_{l}^{2}, ~...~,p_{l}^{N_l})$ are the language instruction tokens from $p_l$. At this stage, the visual and language information are integrated, forming the multi-modal input for the LLM.

More importantly, to endow the LLM with fundamental image understanding capability and realize visual-language alignment, the widely-used natural domain dataset COCO Caption~\cite{chen2015microsoft} is employed for training. To avoid conventional expensive full-parameter fine-tuning and the risk of overfitting, the LoRA\cite{hu2021lora}  technique which is a PEFT approach is adopted in this tuning stage. The LLaMA\cite{touvron2023llama} is utilized as the LLM foundation model. The entire LLaMA weight matrices are frozen during training, and the learnable LoRA rank decomposition matrices are injected into the topmost $L$ layer of the Transformer architecture, greatly decreasing the number of trainable parameters for downstream tasks. The single attention matrix and multi-head attention of  the $l$-th Transformer block can be respectively computed as
\begin{align} 
\mathrm{Att}_l =\mathbf{W} _{Vl}\cdot \mathrm{softmax} \left(\frac{\mathbf{W} _{Ql} (\mathbf{W} _{Kl} )^{T}}{\sqrt{d_k} }\right) ,
\end{align}
\begin{align} 
 \mathrm{MultiAtt}_l\! =\!\!\sum_{h=1}^{H}\mathbf{W} _{Ol}^h \mathbf{W} _{Vl}^h\! \cdot 
\mathrm{softmax} \!\left(\frac{\mathbf{W} _{Ql}^h (\mathbf{W} _{Kl}^h)^{T}}{\sqrt{d_k} }\right)\!,\label{deqn_ex1a}
\end{align} 
where $H $ is the number of attention heads. The $\mathbf{W} _{Ol}^h$, $\mathbf{W} _{Ql}^h$, $\mathbf{W} _{Kl}^h$, and $\mathbf{W} _{Vl}^h\in \mathbb{R} ^{D\times D}$ are weight matrices for each attention head $h \in \mathrm{H}  $ in the $l$-$th$ Transformer block. In particular, four learnable low-rank adapter matrices $\Delta{\mathbf{W} } _{Ql}^h, \Delta{\mathbf{W} } _{Kl}^h$,  $\Delta{\mathbf{W} } _{Vl}^h$, and  $\Delta{\mathbf{W} } _{Ol}^h \in \mathbb{R} ^{D\times D }$ are inserted into the topmost $l$ layers of the Transformer architecture. The adapted multi-head attention is denoted as $\mathrm{Adapted~Attn}$, the output of the $l$-$th$ adapted Transformer attention is defined as
\begin{align}  
 &\mathrm{Adapted~Attn}_l\nonumber\\ 
 &=\sum_{h=1}^{H}\left(\mathbf{W} _{Ol}^h+\Delta{\mathbf{W} } _{Ol}^h  \right)(\mathbf{W} _{Vl}^h+\Delta{\mathbf{W} } _{Vl}^h )\nonumber\\
 &~~\times\!\mathrm{softmax} \!\left(\!\frac{(\mathbf{W} _{Ql}^h\!+\!\Delta{\mathbf{W} } _{Ql}^h ) (\mathbf{W} _{Kl}^h\!+\!\Delta{\mathbf{W} } _{Kl}^h ) ^{T}}{\sqrt{d_k} }\!\right).
\end{align} 

To sum up, this process begins with a frozen LLaMA as the starting point and refines it by optimizing the four smaller learnable matrices, enhancing the mutual understanding between images and language, efficiently transforming language-only LLaMA into a visual-language model.

\subsection{Instruction Adaption Mechanism}
With the methods documented in Section IV-A, a suitable visual-language alignment is achieved. 
To further adapt to multi-source and multi-granularity detection ship tasks, the visual-language model is continued fine-tuning on the newly constructed MMShip dataset. 

To release the cross-modal learning potential of LLM and to augment the instruction-following ability, the instruction adapter mechanism is developed. As illustrated in Fig. \ref{overall} (c), more learnable parameters are added in the instruction adaption stage, compared to the visual-language alignment one. Specifically, for each linear layer in the Transformer, a bias matrix $\Delta {{\bf W}_b}$ and a scale $\Delta {{\bf W}_s}$ factors are inserted as two trainable parameters. Given a linear layer $f(x) =\mathbf{W} x$, it can be transformed into equation as 
\begin{equation}
\label{deqn_ex1a}
f(x) = \Delta {\mathbf{W} _s} (\mathbf{W}x +\Delta {\mathbf{W} _b} ),
\end{equation}
with learnable matrices $\mathbf{W}, \Delta {\mathbf{W} _b},$ and $ \Delta {\mathbf{W} _s} \in \mathbb{R}^{D\times D} $. The added two parameters are initialized with zeros and a random Gaussian, respectively, keeping fine-tuning stability and effectiveness. The mathematical formula can be expressed as $\Delta {\mathbf{W} _b}=\mathrm{Init} (0), \Delta {\mathbf{W} _s}  \sim \mathcal{N}(\mu, \sigma^2)$.

Note that these added parameters only account extremely tiny fraction of the entire model, ensuring that Popeye remains lightweight training. The stage-wise parameter optimization method effectively addresses the challenges of interference between image-text mutual understanding and task-specific instruction-following for ship detection in the RS domain, thereby enhancing the emergent ability of Popeye to follow instructions to handle multi-source and multi-granularity ship detection tasks interactively. Furthermore, by utilizing the designed cross-modal joint training strategy, Popeye is equipped with robust zero-shot reasoning ability.

\subsection{The Integration with SAM}
In addition to ship detection capability, we also integrate the proposed Popeye with SAM\cite{kirillov2023segment} to tackle the more challenging language-referred pixel-level segmentation task. SAM is an open-ended image segmentation model that allows for promptable segmentation. However, ship images from RS imagery contain complex background interference and vague object edges, which affect SAM's segmentation efficacy in this domain. The integration of our model with SAM enhances the capabilities of SAM specifically in the context of ship RS images.  As illustrated in Fig. \ref{example}, our Popeye can generate accurate ship HBB detection results based on the language instructions, and these results can be regarded as the prior prompts for SAM. Specifically, the SAM encoder uses the Masked Autoencoder (MAE)~\cite{he2022masked} to encode high-resolution image inputs into visual features and the bounding boxes generated from Popeye into the prompts embedding tokens. Then, the mask decoder efficiently enhances the interaction between image features and prompt embeddings, facilitating the generation of the mask output. The overall process\cite{chen2023rsprompter} can be expressed as 
\begin{equation}
F_\mathrm{visual} =\mathrm{SAM } _\mathrm{v-enc}(I),
\end{equation}
\begin{equation}
B_\mathrm{det} =\mathrm{Popeye }(I),
\end{equation}
\begin{equation}
F_\mathrm{prompt} =\mathrm{SAM } _\mathrm{p-enc}(B_\mathrm{det}),
\end{equation}
\begin{equation}
\Omega    =\mathrm{SAM } _\mathrm{dec}(F_\mathrm{visual},F_\mathrm{prompt}),
\end{equation}
where $I \in \mathbb{R} ^{H\times W\times 3}$ represents the input image, $F_\mathrm{visual}$ refers to the visual features extracted by the SAM image encoder. $B_\mathrm{det}$ represents the sparse prompts including ship detection bounding boxes, $F_\mathrm{prompt}$ represents the sparse embedding tokens encoded by the prompt encoder and $\Omega $ is the set of predicted masks. 

In summary, the integration of Popeye with SAM enables an expansion of the language-guided ship segmentation at the pixel level, without additional training expenses. The proposed visual-language model Popeye can effectively unify ship HBB detection, ship OBB detection, and ship pixel-level segmentation tasks, Moreover, Popeye allows users to retrieve ship targets in RS images in a language-interactive manner. Fig. \ref{example}. shows examples of the applications for ship language-referred segmentation.

\section{Experiments}
In this section, extensive experiments are conducted to validate the performance of the proposed Popeye. Firstly, we detail the construction method of the ship instruction training dataset in Section IV-A. Subsequently, we present the implementation details in Section IV-B. Furthermore, the ship detection, comparisons between Popeye and other VLMs, and segmentation experimental results are addressed in Sections IV-C, IV-D, and IV-E, respectively.

\begin{figure*}[!t]
	\centering
	\includegraphics[width=6.5in]{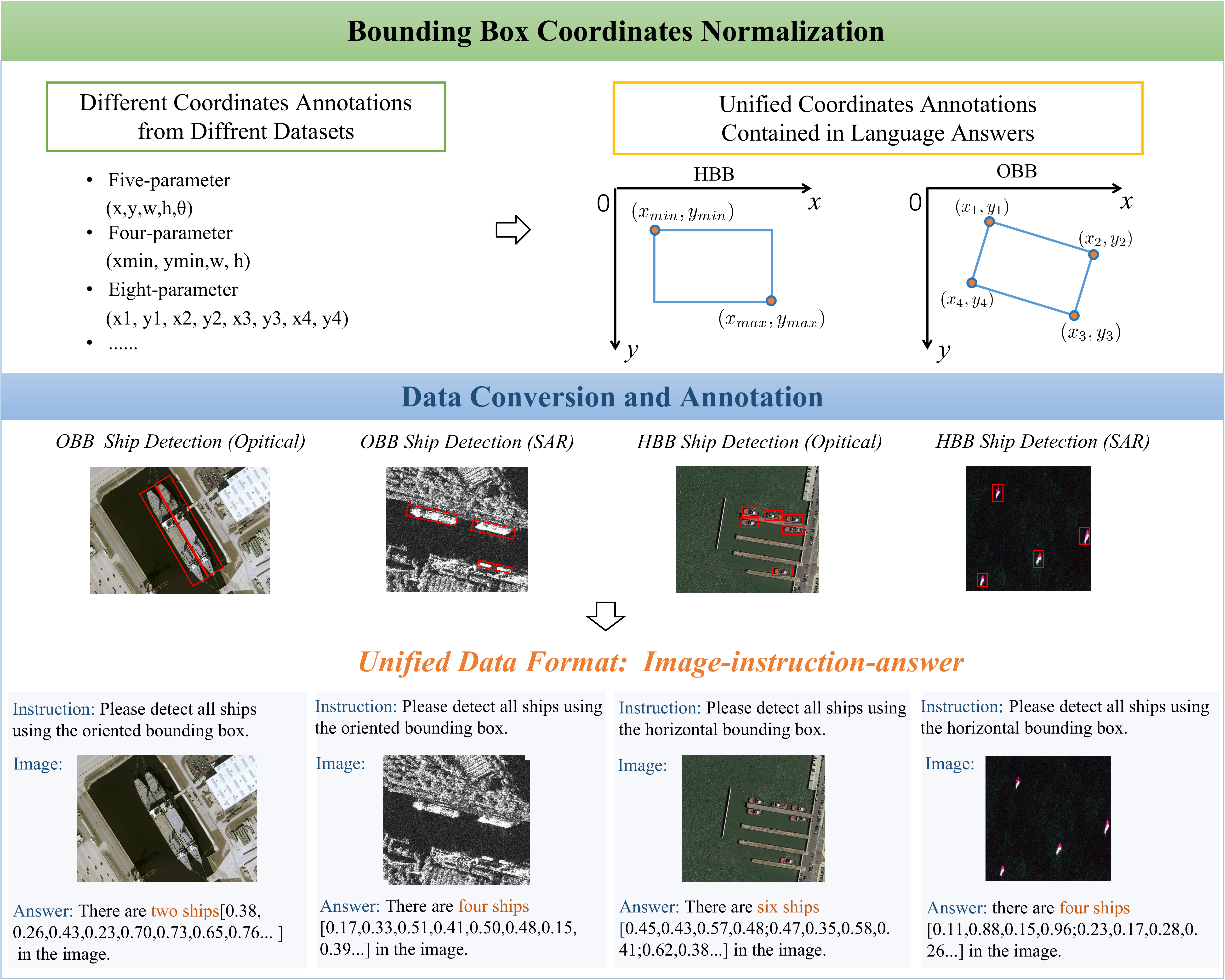}
	\caption{Constructing the MMShip dataset follows the unified labeling paradigm: transforming and annotating the existing multi-source ship detection data into uniform image-instruction-answer format.}
	\label{FIG:dataset}
\end{figure*}

\subsection{MMShip Dataset Construction}
To ensure compatibility across various remote sensing visual modalities and detection tasks within Popeye, a unified labeling paradigm is developed. We utilize existing ship object detection datasets as the foundational data to construct a new dataset named MMShip, featuring multi-source multi-modal instruction-following. The MMShip dataset contains 81k instruction data of high-quality ship images. Specifically, four main ship detection datasets are transformed and annotated, namely dataset DOSR~\cite{han2021fine} DOTA ship subset~\cite{9560031}, SSDD~\cite{zhang2021sar}, and HRSID~\cite{wei2020hrsid}, which are annotated by different formats including four-parameter, five-parameter, and eight-parameter method. We aim to convert and re-annotate the different data formats to a uniform image-instruction-answer format. As shown in the upper of Fig. \ref{FIG:dataset}, the bounding box coordinates undergo a normalization process, and then the answers include unified coordinates annotations. In simpler terms, HBB is defined by two points' coordinates, whereas OBB requires four points' coordinates to locate. The multi-source data conversion and annotation process is detailed as follows.\\

To construct conversation data containing image-instruction-answer pairings, specific language instructions are employed to guide the model towards predicting either HBB or OBB. For instance, for HBB detection, the instruction is ``\textit{Please detect all ships using the horizontal bounding box.}''. Similarly, for OBB detection, the instruction is ``\textit{Please detect all ships using the oriented bounding box.}''. Then, the formats of the answer are as follows: in the HBB format, a bounding box is defined by the coordinates \([x_{\text{min}}, y_{\text{min}}, x_{\text{max}}, y_{\text{max}}]\). The points \((x_{\text{min}}, y_{\text{min}})\) and \((x_{\text{max}}, y_{\text{max}})\) are identified as the corners of the bounding box that are, closest to and farthest from the origin of the coordinate system, respectively. Conversely, the OBB format is specified as \([x_1, y_1, x_2, y_2, x_3, y_3, x_4, y_4]\). Within this format, the point \((x_1, y_1)\) is designated as the corner of the bounding box nearest to the coordinate origin, with the subsequent points \((x_2, y_2)\), \((x_3, y_3)\), and \((x_4, y_4)\) arranged in a clockwise order. As shown in Fig. \ref{FIG:dataset}, utilizing the designed unified labeling paradigm, the HBB and OBB multi-source detection data is transformed into the uniform image-instruction-answer format, as well as the answers contain object coordinates information. 
\begin{figure*}[!h]
\centering
\includegraphics[width=6.5in]{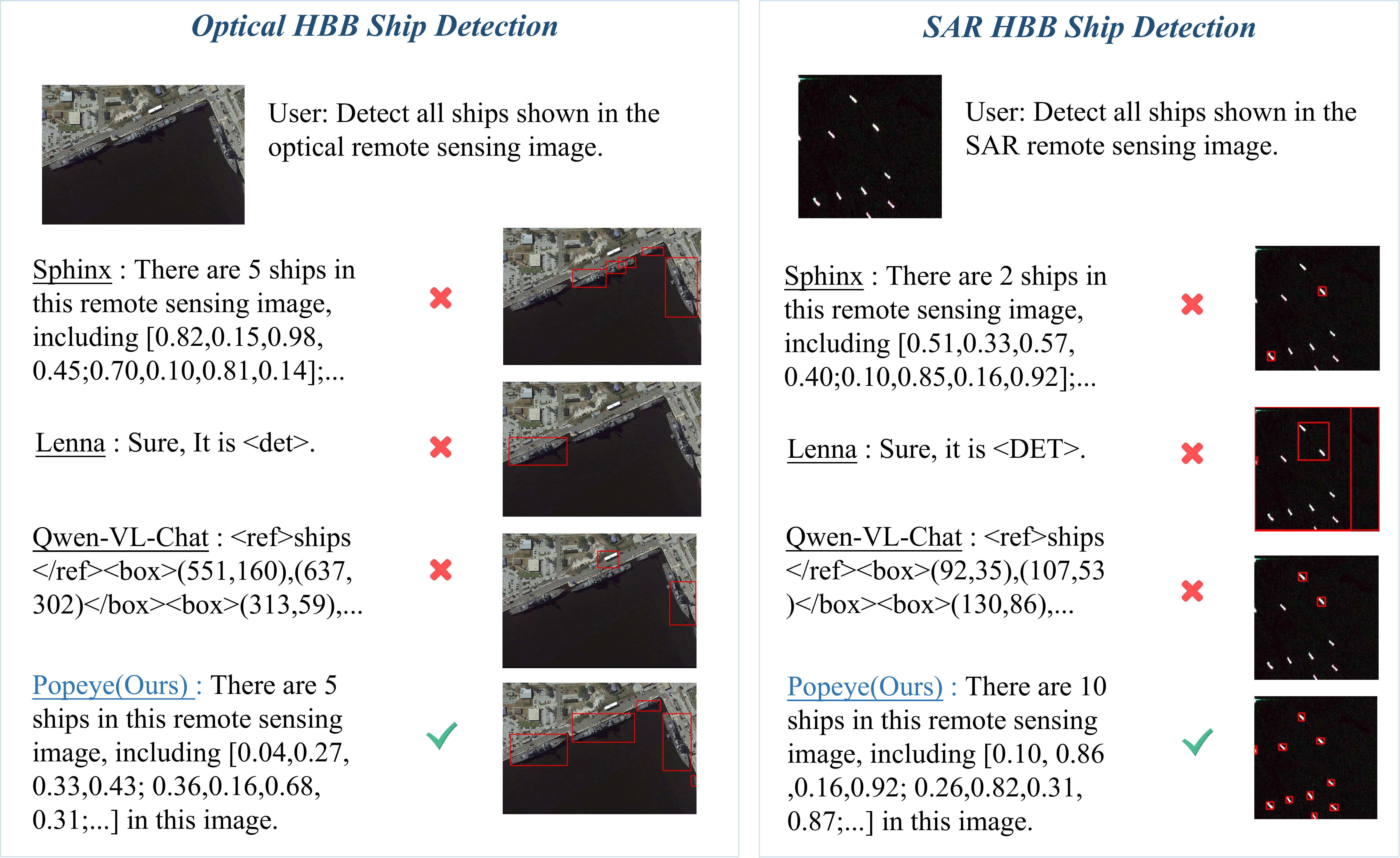}
\caption{Examples for performance comparison between Popeye and other VLMs.}
\label{compare}
\end{figure*}

\subsection{Implementation Details}

In this part, we present the training configuration. Particularly, in the visual-language alignment stage, we start with the off-the-shelf open-source weights LLaMA-7B~\cite{touvron2023llama} with 32 transformer layers and insert trainable LoRA matrices into the last $L= 30 $ transformers layers. The multiple visual encoders are kept frozen during the training. The visual projections are initialized randomly. We train our model using the AdamW optimizer~\cite{kingma2014adam} with a cosine learning rate scheduler. In the ship domain adaption stage,  the optimizer settings are similar to the Pre-training stage. The trained and frozen parts remain the same as the visual-language alignment stage except the newly added bias and scale of the linear layer are learnable.

\subsection{Ship Object Detection}
The evaluation of ship object detection is divided into HBB detection and OBB detection. To verify the potential of our proposed Popeye model in ship object HBB detection, we employ the zero-shot setting and compare Popeye with other VLMs and Open-vocabulary object detection models on three datasets including two optical ship detection datasets ShipRSimagenet, HRSC2016, and a SAR ship detection dataset DSSDD.
To address the challenge of VLMs not predicting confidence scores, we employ clip-score as a confidence logit. Remoteclip~\cite{liu2023remoteclip} weights are adopted to compute the clipscore.
For OBB detection,  we adopt the OBB format of HRSC2016, DSSDD for zero-shot comparison with specialist models trained on DOTA. Additionally, the experiments mentioned above are conducted on the test set of the corresponding dataset.

As depicted in Tab. \ref{tab:ShipRSImageNet_compare_zeroshot}, for HBB detection, Popeye exhibits notable improvements of 7.41$\%$, 7.89$\%$, 9.02$\%$ on AP@40, AP@50, AP@60 when contrasted with other VLMs and Open-vocabulary models on ShipRSimagenet. For the DSSDD dataset, when conducting HBB detection, Popeye brings improvements of 40.55$\%$, 26.18$\%$, 4.16$\%$ on AP@40, AP@50, and AP@60, compared with VLM model Sphinx, which is shown in Tab. \ref{tab:DSSDD_compare_zeroshot}. On the HRSC2016 dataset, Popeye suppresses the VLM Lenna with remarkable improvements of 13.31$\%$, 7.35$\%$, 6.51$\%$ on AP@40, AP@50 and AP@60. Comparative results on HRSC2016 are demonstrated on Tab. \ref{tab:HRSC2016_compare_zeroshot}. For OBB detection, it is observed from Tab. \ref{tab:DSSDD_compare_zeroshot_obb} that Popeye surpasses other specialist models trained on DOTA with the improvements of 4.63$\%$ and 2.86$\%$ on AP@40 and AP@50. Tab. \ref{tab:HRSC2016_compare_zeroshot_obb} also verifies that Popeye has a higher detection accuracy on HRSC2016 compared to other expert methods. These results demonstrate the potential and superiority of Popeye in ship target detection and also prove its powerful generalization ability in completely new, unseen environments. Furthermore, Fig. \ref{example2} provides a clear visualization of Popeye's superior zero-shot detection accuracy in HBB and OBB formats.

\begin{table}[!htbp]
\caption{Zero-shot comparison results for HBB detection on ShipRSImageNet for other methods and our method Popeye.}\label{tab:ShipRSImageNet_compare_zeroshot}
\renewcommand{\arraystretch}{1.5}
\scalebox{0.9}{
\begin{tabular}{lccccc}  
\toprule      
\multicolumn{1}{c}{Method}          & \multicolumn{1}{c|}{Publication Year} & AP@40 & AP@50  & AP@60\\ 
\cmidrule(lr){1-1}\cmidrule(lr){2-5}
\textit{\textbf{Open-vocabulary Model}}        &    &       &       &     \\ 

\multicolumn{1}{l}{GroundingDINO~\cite{liu2023grounding}} &  \multicolumn{1}{l|}{Arxiv 2023}       & 35.13 & 33.27        & 30.96           \\
\multicolumn{1}{l}{mm-GroundingDINO~\cite{zhao2024open}}    & \multicolumn{1}{l|}{Arxiv 2024}     & 38.19    & 37.18     &   35.57       \\
\cmidrule(lr){1-1}\cmidrule(lr){2-5}
\textit{\textbf{Visual-language Model}}        &    &       &       &     \\ 

\multicolumn{1}{l}{Lenna~\cite{wei2023lenna}}      & \multicolumn{1}{l|}{Arxiv 2023}       & 49.27   &47.41   & 44.51    \\
\multicolumn{1}{l}{Qwen-VL-Chat~\cite{bai2023qwen}}      & \multicolumn{1}{l|}{Arxiv 2023}       & 30.34   &22.69  &  21.32    \\
\multicolumn{1}{l}{Sphinx~\cite{lin2023sphinx}}      & \multicolumn{1}{l|}{Arxiv 2023}       & 41.61  &40.82     & 40.70       \\
\cmidrule(lr){1-1}\cmidrule(lr){2-5}
\rowcolor[gray]{0.95} \multicolumn{1}{l}{\cellcolor[gray]{0.95}\textbf{Popeye(Ours)}}                  & \multicolumn{1}{l|}{}    & \textbf{56.68 } & \textbf{55.30}  & \textbf{53.53}           \\ 
\bottomrule
\end{tabular}
} 
\end{table}

\begin{table}[!htbp]
\caption{Zero-shot comparison results for HBB detection on DSSDD for other methods and our method Popeye.}\label{tab:DSSDD_compare_zeroshot}
\renewcommand{\arraystretch}{1.5}
\scalebox{0.9}{
\begin{tabular}{lcccc}  
\toprule      
\multicolumn{1}{c}{Method}          & \multicolumn{1}{c|}{Publication Year} & AP@40 & AP@50  &AP@60\\ 
\cmidrule(lr){1-1}\cmidrule(lr){2-5}
\textit{\textbf{Open-vocabulary Model}}        &    &       &       &     \\ 
\multicolumn{1}{l}
{GroundingDINO~\cite{liu2023grounding}} &  \multicolumn{1}{l|}{Arxiv 2023}       & 22.31 & 20.24         &  14.68          \\
\multicolumn{1}{l}{mm-GroundingDINO~\cite{zhao2024open}}    & \multicolumn{1}{l|}{Arxiv 2024}     & 7.76    & 6.63      &  3.99        \\
\cmidrule(lr){1-1}\cmidrule(lr){2-5}
\textit{\textbf{Visual-language Model}}        &    &       &       &     \\ 
\multicolumn{1}{l}{Lenna~\cite{wei2023lenna}}      & \multicolumn{1}{l|}{Arxiv 2023}        & 11.79   &9.89   &  5.84   \\
\multicolumn{1}{l}{Qwen-VL-Chat~\cite{bai2023qwen}}      & \multicolumn{1}{l|}{Arxiv 2023}       & 17.16   &16.16   & 11.42     \\
\multicolumn{1}{l}{Sphinx~\cite{lin2023sphinx}}      & \multicolumn{1}{l|}{Arxiv 2023}       &35.04  &25.69     & 22.29      \\
\cmidrule(lr){1-1}\cmidrule(lr){2-5}
\rowcolor[gray]{0.95} \multicolumn{1}{l}{\cellcolor[gray]{0.95}\textbf{Popeye(Ours)}}                  & \multicolumn{1}{l|}{}    & \textbf{75.59}  & \textbf{51.87}  & \textbf{26.45}           \\ 
\bottomrule
\end{tabular}
} 
\end{table}

\begin{table}[!htbp]
\caption{Zero-shot comparison results for HBB detection on HRSC2016 for other methods and our method Popeye.}\label{tab:HRSC2016_compare_zeroshot}
\renewcommand{\arraystretch}{1.5}
\scalebox{0.9}{
\begin{tabular}{lcccc}  
\toprule      
\multicolumn{1}{c}{Method}          & \multicolumn{1}{c|}{Publication Year} & AP@40 & AP@50  & AP@60\\ 
\cmidrule(lr){1-1}\cmidrule(lr){2-5}
\textit{\textbf{Open-vocabulary Model}}        &    &       &       &     \\ 

\multicolumn{1}{l}{GroundingDINO~\cite{liu2023grounding}} &  \multicolumn{1}{l|}{Arxiv 2023}       & 43.14 & 40.71      &    37.92        \\
\multicolumn{1}{l}{mm-GroundingDINO~\cite{zhao2024open}}    & \multicolumn{1}{l|}{Arxiv 2024}     & 48.47    & 47.59     &  46.12        \\
\cmidrule(lr){1-1}\cmidrule(lr){2-5}
\textit{\textbf{Visual-language Model}}        &    &       &       &     \\ 

\multicolumn{1}{l}{Lenna~\cite{wei2023lenna}}      & \multicolumn{1}{l|}{Arxiv 2023}       & 57.05   &55.14   &52.05   \\
\multicolumn{1}{l}{Qwen-VL-Chat~\cite{bai2023qwen}}      & \multicolumn{1}{l|}{Arxiv 2023}       & 39.23   &30.21  & 22.39     \\
\multicolumn{1}{l}{Sphinx~\cite{lin2023sphinx}}      & \multicolumn{1}{l|}{Arxiv 2023}       & 56.11  &55.32     & 54.38      \\
\cmidrule(lr){1-1}\cmidrule(lr){2-5}
\rowcolor[gray]{0.95} \multicolumn{1}{l}{\cellcolor[gray]{0.95}\textbf{Popeye(Ours)}}                  & \multicolumn{1}{l|}{}    & \textbf{70.36}  & \textbf{62.67 } & \textbf{60.89 }           \\ 
\bottomrule
\end{tabular}
} 
\end{table}

\begin{table}[!htbp]
\caption{Zero-shot comparison results for OBB detection on HRSC2016 for other methods and our method Popeye.}\label{tab:HRSC2016_compare_zeroshot_obb}
\renewcommand{\arraystretch}{1.5}
\scalebox{0.9}{
\begin{tabular}{lcccc}  
\toprule      
\multicolumn{1}{c}{Method}          & \multicolumn{1}{c|}{Publication Year} & AP@40 & AP@50 &AP@60 \\ 
\cmidrule(lr){1-1}\cmidrule(lr){2-5}
\textit{\textbf{Specialist Model}}        &    &       &       &      \\ 
\multicolumn{1}{l}{S2A-Net~\cite{han2021align}}      & \multicolumn{1}{l|}{TGRS 2021}       & 47.03   &42.87 &     37.30  \\
\multicolumn{1}{l}{CFA~\cite{guo2021beyond}}      & \multicolumn{1}{l|}{CVPR 2021}       & 47.62  &44.97  & 39.55         \\
\multicolumn{1}{l}{Oriented RepPoints~\cite{li2022oriented}}      & \multicolumn{1}{l|}{CVPR 2022}       & 35.23  &31.87   &27.39       \\
\multicolumn{1}{l}{Oriented R-CNN~\cite{xie2021oriented}}      & \multicolumn{1}{l|}{ICCV 2021}       & 52.99  &51.79   &  \textbf{47.17}       \\
\multicolumn{1}{l}{Sasm~\cite{hou2022shape}}      & \multicolumn{1}{l|}{AAAI 2022}       & 31.45  &28.27   & 25.14        \\
\multicolumn{1}{l}{R3Det~\cite{yang2021r3det}}      & \multicolumn{1}{l|}{AAAI 2021}       & 36.11  &30.76      &24.20     \\

\cmidrule(lr){1-1}\cmidrule(lr){2-5}
\textit{\textbf{Visual-language Model}}        &    &       &       &      \\ 
\rowcolor[gray]{0.95} \multicolumn{1}{l}{\cellcolor[gray]{0.95}\textbf{Popeye(Ours)}}                  & \multicolumn{1}{l|}{}    & \textbf{58.15}  & \textbf{54.91}     & 44.94       \\ 
\bottomrule
\end{tabular}
} 
\end{table}

\begin{table}[!!htbp]
\caption{Zero-shot OBB detection comparison results for OBB detection on DSSDD for other methods and our method Popeye.}\label{tab:DSSDD_compare_zeroshot_obb}
\renewcommand{\arraystretch}{1.5}
\scalebox{0.9}{
\begin{tabular}{lcccc}  
\toprule      
\multicolumn{1}{c}{Method}          & \multicolumn{1}{c|}{Publication Year} & AP@40 & AP@50 & AP@60\\ 
\cmidrule(lr){1-1}\cmidrule(lr){2-5}
\textit{\textbf{Specialist Model}}        &    &       &       &      \\ 
\multicolumn{1}{l}{S2A-Net~\cite{han2021align}}      & \multicolumn{1}{l|}{TGRS 2021}       & 37.44   &35.10 &25.79       \\
\multicolumn{1}{l}{CFA~\cite{guo2021beyond}}      & \multicolumn{1}{l|}{CVPR 2021}       & 30.44  &24.62  & 16.76         \\
\multicolumn{1}{l}{Oriented RepPoints~\cite{li2022oriented}}      & \multicolumn{1}{l|}{CVPR 2022}       & 22.61  &20.14   &15.74         \\
\multicolumn{1}{l}{Oriented R-CNN~\cite{xie2021oriented}}      & \multicolumn{1}{l|}{ICCV 2021}       & 17.34  &16.92   & 15.81        \\
\multicolumn{1}{l}{R3Det~\cite{yang2021r3det}}      & \multicolumn{1}{l|}{AAAI 2021}       & 58.49  &49.05      & 30.35     \\

\cmidrule(lr){1-1}\cmidrule(lr){2-5}
\textit{\textbf{Visual-language Model}}        &    &       &       &      \\ 
\rowcolor[gray]{0.95} \multicolumn{1}{l}{\cellcolor[gray]{0.95}\textbf{Popeye(Ours)}}                  & \multicolumn{1}{l|}{}    & \textbf{63.12}  & \textbf{51.92}      &  \textbf{38.35 }     \\ 
\bottomrule
\end{tabular}
} 
\end{table}
\subsection{Comparison Between Popeye and Other VLMs}
In this section, we conduct a comparative analysis of Popeye's performance against existing natural scene VLMs when confronted with challenging samples in the ship domain. Note that existing nature scenes VLMs are incapable of OBB detection due to the absence of OBB format datasets. Therefore, we choose the HBB object detection task and use the zero-shot setting to demonstrate Popeye’s superior performance. As shown in the left part of Fig. \ref{compare}, when facing an optical image featuring multiple nearshore ships from the HRSC dataset, VLMs like Lenna~\cite{wei2023lenna}, Sphinx~\cite{lin2023sphinx}, and Qwen-VL-Chat~\cite{bai2023qwen} generate missed detection and false alarms due to the interference of surrounding objects and blurry maritime environment. In contrast, Popeye accurately detects all the ships, even if there is a ship where only the bow is visible. Furthermore, when facing a SAR detection scenario from the DSSDD dataset (the right part of Fig. \ref{compare}), VLMs like Lenna, Qwen-VLChat, and Sphinx all struggle to recognize all the ships accurately. Note that Popeye has better visual perception for small-scale targets and can correctly detect the ships in off-shore SAR imagery.

\subsection{Segmentation Visualization}
As introduced in Section III-E, Popeye is integrated with the off-the-shelf SAM, seamlessly obtaining pixel-level segmentation ability. Popeye generates the horizontal bounding boxes for the ship target and serves as the prior prompts for SAM to achieve ship instance segmentation. To evaluate Popeye's segmentation performance, we select more challenging RS images of ships, including optical and SAR images. These images are blurry to recognition, or the ships are camouflaged by the complex background or the ship targets are very tiny. As shown in Fig. \ref{example2}, in all these tested cases, we observe that the predicted segmentation masks generated by Popeye tiny and blurred ships are extremely accurate. 

Furthermore, we clarify the distinctions between Popeye and SAM. Popeye is a visual-language model tailored for ship interpretation, capable of completing various tasks through natural language prompts in multi-turn dialogues. In contrast, SAM is a fundamental segmentation model that can perform segmentation tasks through interactive inputs, including mouse clicks (points) and drawing the boxes. Notably, Popeye can accomplish both language-referring (e.g., ``Detect ships") object detection and segmentation, whereas class-agnostic SAM is limited to segmentation tasks. In addition, SAM can achieve pixel-level segment everything but lacks semantic perception, while Popeye primarily focuses on ship object segmentation.

\section{Discussion}

\textbf{Review.} In this paper, the various ship detection tasks and multi-source imagery are comprehended in one architecture effectively. This ability is owing to the proposed unified learning framework and the uniform data format. Notably, the experiments have demonstrated that Popeye performs well in zero-shot ship detection. The zero-shot capability is primarily attributed to the design of a cross-domain joint training structure and data diversity. This design equips the model with extensive knowledge of natural imagery and facilitates it to develop open-vocabulary reasoning. Most importantly, we utilize robust hybrid visual experts to realize enhanced visual perception, based on an off-the-shelf LLM and develop it into a VLM for ship detection in the RS domain. It is shown that leveraging the foundational model and maximizing its advantages in specific tasks within a particular domain such as ship is an effective and efficient approach.

\textbf{Impacts.} The proposed Popeye fills a significant gap in the development of VLMs within the maritime domain. Popeye establishes a visual-language foundational framework for further more sophisticated applications in maritime monitoring and management. Most importantly, Popeye can perform multiple analyses through language interaction in multi-turn conversations. We anticipate that this interactive mode can revolutionize how users engage with the monitoring system and the ship interpretation workflows. By advancing towards a multi-task learning framework, Popeye could simultaneously perform multiple analysis types in real-world scenarios, making it a more robust tool for maritime surveillance.

\textbf{Limitations.} The proposed Popeye is the first VLM tailored for multi-granularity ship detection in the RS domain. Nevertheless, there still exist limitations. For example, this paper primarily discusses ship detection, it doesn't incorporate other ship-related tasks, such as ship classification, or more fine-grained tasks like visual grounding, which could provide more detailed information and the ship's immediate surroundings. Moreover, infrared imagery is important under various conditions (e.g., night, fog). However, the infrared modality is not integrated into the proposed framework at present, restricting its applicability and scope.

\textbf{Future work.} In the future, we plan to incorporate a broader range of ship visual tasks and more visual modalities into Popeye, aiming to enhance its all-purpose capabilities. For example, future versions of Popeye could include capabilities for classifying ships such as warships, tankers, or passenger vessels. This enhancement would enable users to gain insights into the potential activities in observed maritime regions. Moreover, developing visual grounding ability would allow Popeye to link specific textual descriptions to precise image regions, enhancing its utility for detailed visual inspections and situational awareness. Furthermore, we plan to expand Popeye’s framework to include infrared and other spectral bands, improving model performance in challenging visibility conditions.

\section{Conclusion}
In this paper, a unified visual-language model called Popeye has been proposed, excelling in uniformly handling multi-granularity ship detection tasks like HBB, OBB, and pixel-level ship segmentation. Technically, a unified labeling paradigm has been developed to construct a dataset called MMShip, containing 81k multi-modal ship instruction-following data and covering multi-source RS images such as SAR and optical. Subsequently, a cross-modal image interpretation method and instruction adaption mechanism for the ship RS domain have been constructed, leveraging the language as a medium for bridging visual and language contexts and realizing a more universal paradigm for multi-source ship interpretation. In addition, Popeye is integrated with SAM to extend instance segmentation functionality without extra training expenses. Furthermore, extensive experiments have demonstrated that Popeye achieves robust zero-shot performances in multi-source ship imagery HBB detection, OBB detection, and pixel-level segmentation through natural language interactions.

\bibliographystyle{unsrt}
\bibliography{ref.bib}

\begin{thebibliography}{10}

\bibitem{han2021shipyolo}
Xu~Han, Lining Zhao, Yue Ning, and Jingfeng Hu.
\newblock Shipyolo: an enhanced model for ship detection.
\newblock {\em Journal of Advanced Transportation}, 2021:1--11, 2021.

\bibitem{cui2019dense}
Zongyong Cui, Qi~Li, Zongjie Cao, and Nengyuan Liu.
\newblock Dense attention pyramid networks for multi-scale ship detection in sar images.
\newblock 57(11):8983--8997, 2019.

\bibitem{10547418}
Wei Zhang, Miaoxin Cai, Tong Zhang, Yin Zhuang, and Xuerui Mao.
\newblock Earthgpt: A universal multi-modal large language model for multi-sensor image comprehension in remote sensing domain.
\newblock {\em IEEE Transactions on Geoscience and Remote Sensing}, 2024, early access.

\bibitem{zhu2010novel}
Changren Zhu, Hui Zhou, Runsheng Wang, and Jun Guo.
\newblock A novel hierarchical method of ship detection from spaceborne optical image based on shape and texture features.
\newblock {\em IEEE transactions on geoscience and remote sensing}, 48(9):3446--3456, 2010.

\bibitem{wang2021review}
Liqian Wang, Shuzhen Fan, Yunxia Liu, Yongfu Li, Cheng Fei, Junliang Liu, Bohan Liu, Yakui Dong, Zhaojun Liu, and Xian Zhao.
\newblock A review of methods for ship detection with electro-optical images in marine environments.
\newblock {\em Journal of Marine Science and Engineering}, 9(12):1408, 2021.

\bibitem{dong2018ship}
Chao Dong, Jinghong Liu, and Fang Xu.
\newblock Ship detection in optical remote sensing images based on saliency and a rotation-invariant descriptor.
\newblock {\em Remote Sensing}, 10(3):400, 2018.

\bibitem{zhang2020intelligent}
Yulian Zhang, Lihong Guo, Zengfa Wang, Yang Yu, Xinwei Liu, and Fang Xu.
\newblock Intelligent ship detection in remote sensing images based on multi-layer convolutional feature fusion.
\newblock {\em Remote Sensing}, 12(20):3316, 2020.

\bibitem{li2021domain}
Linhao Li, Zhiqiang Zhou, Bo~Wang, Lingjuan Miao, Zhe An, and Xiaowu Xiao.
\newblock Domain adaptive ship detection in optical remote sensing images.
\newblock {\em Remote Sensing}, 13(16):3168, 2021.

\bibitem{liu2017rotated}
Zikun Liu, Jingao Hu, Lubin Weng, and Yiping Yang.
\newblock Rotated region based cnn for ship detection.
\newblock In {\em 2017 IEEE International Conference on Image Processing (ICIP)}, pages 900--904. IEEE, 2017.

\bibitem{li2020novel}
Linhao Li, Zhiqiang Zhou, Bo~Wang, Lingjuan Miao, and Hua Zong.
\newblock A novel cnn-based method for accurate ship detection in hr optical remote sensing images via rotated bounding box.
\newblock {\em IEEE Transactions on Geoscience and Remote Sensing}, 59(1):686--699, 2020.

\bibitem{sun2021anchor}
Zhongzhen Sun, Muchen Dai, Xiangguang Leng, Yu~Lei, Boli Xiong, Kefeng Ji, and Gangyao Kuang.
\newblock An anchor-free detection method for ship targets in high-resolution sar images.
\newblock {\em IEEE Journal of Selected Topics in Applied Earth Observations and Remote Sensing}, 14:7799--7816, 2021.

\bibitem{chen2020sar}
Yuan Chen, Jie Yu, and Yang Xu.
\newblock Sar ship target detection for ssdv2 under complex backgrounds.
\newblock In {\em 2020 International Conference on Computer Vision, Image and Deep Learning (CVIDL)}, pages 560--565. IEEE, 2020.

\bibitem{zhu2020rapid}
Mingming Zhu, Guoping Hu, Hao Zhou, Chunguang Lu, Yule Zhang, Shijie Yue, and Yao Li.
\newblock Rapid ship detection in sar images based on yolov3.
\newblock In {\em 2020 5th international conference on communication, image and signal processing (CCISP)}, pages 214--218. IEEE, 2020.

\bibitem{zhao2020attention}
Yan Zhao, Lingjun Zhao, Boli Xiong, and Gangyao Kuang.
\newblock Attention receptive pyramid network for ship detection in sar images.
\newblock {\em IEEE Journal of Selected Topics in Applied Earth Observations and Remote Sensing}, 13:2738--2756, 2020.

\bibitem{chang2019ship}
Yang-Lang Chang, Amare Anagaw, Lena Chang, Yi~Chun Wang, Chih-Yu Hsiao, and Wei-Hong Lee.
\newblock Ship detection based on yolov2 for sar imagery.
\newblock {\em Remote Sensing}, 11(7):786, 2019.

\bibitem{he2021learning}
Yishan He, Fei Gao, Jun Wang, Amir Hussain, Erfu Yang, and Huiyu Zhou.
\newblock Learning polar encodings for arbitrary-oriented ship detection in sar images.
\newblock {\em IEEE Journal of Selected Topics in Applied Earth Observations and Remote Sensing}, 14:3846--3859, 2021.

\bibitem{zhao2021anchor}
Xin Zhao, Bo~Zhang, Zhixin Tian, Changgui Xu, Fan Wu, and Chunling Sun.
\newblock An anchor-free method for arbitrary-oriented ship detection in sar images.
\newblock In {\em 2021 SAR in Big Data Era (BIGSARDATA)}, pages 1--4. IEEE, 2021.

\bibitem{brown2020language}
Tom Brown, Benjamin Mann, Nick Ryder, Melanie Subbiah, Jared~D Kaplan, Prafulla Dhariwal, Arvind Neelakantan, Pranav Shyam, Girish Sastry, Amanda Askell, et~al.
\newblock Language models are few-shot learners.
\newblock {\em Advances in neural information processing systems}, 33:1877--1901, 2020.

\bibitem{touvron2023llama}
Hugo Touvron, Thibaut Lavril, Gautier Izacard, Xavier Martinet, Marie-Anne Lachaux, Timoth{\'e}e Lacroix, Baptiste Rozi{\`e}re, Naman Goyal, Eric Hambro, Faisal Azhar, et~al.
\newblock Llama: Open and efficient foundation language models.
\newblock {\em arXiv preprint arXiv:2302.13971}, 2023.

\bibitem{zhang2022opt}
Susan Zhang, Stephen Roller, Naman Goyal, Mikel Artetxe, Moya Chen, Shuohui Chen, Christopher Dewan, Mona Diab, Xian Li, Xi~Victoria Lin, et~al.
\newblock Opt: Open pre-trained transformer language models.
\newblock {\em arXiv preprint arXiv:2205.01068}, 2022.

\bibitem{openai2023chatgpt}
OpenAI.
\newblock Chatgpt.
\newblock \url{https://chat.openai.com}, 2023a.

\bibitem{openai2023gpt4}
OpenAI.
\newblock Gpt-4 technical report, 2023.

\bibitem{radford2019language}
Alec Radford, Jeffrey Wu, Rewon Child, David Luan, Dario Amodei, Ilya Sutskever, et~al.
\newblock Language models are unsupervised multitask learners.
\newblock {\em OpenAI blog}, 1(8):9, 2019.

\bibitem{zhu2023minigpt}
Deyao Zhu, Jun Chen, Xiaoqian Shen, Xiang Li, and Mohamed Elhoseiny.
\newblock Minigpt-4: Enhancing vision-language understanding with advanced large language models.
\newblock {\em arXiv preprint arXiv:2304.10592}, 2023.

\bibitem{liu2023visual}
Haotian Liu, Chunyuan Li, Qingyang Wu, and Yong~Jae Lee.
\newblock Visual instruction tuning.
\newblock {\em arXiv preprint arXiv:2304.08485}, 2023.

\bibitem{zhang2023llama}
Renrui Zhang, Jiaming Han, Aojun Zhou, Xiangfei Hu, Shilin Yan, Pan Lu, Hongsheng Li, Peng Gao, and Yu~Qiao.
\newblock Llama-adapter: Efficient fine-tuning of language models with zero-init attention.
\newblock {\em arXiv preprint arXiv:2303.16199}, 2023.

\bibitem{gao2023llama}
Peng Gao, Jiaming Han, Renrui Zhang, Ziyi Lin, Shijie Geng, Aojun Zhou, Wei Zhang, Pan Lu, Conghui He, Xiangyu Yue, et~al.
\newblock Llama-adapter v2: Parameter-efficient visual instruction model.
\newblock {\em arXiv preprint arXiv:2304.15010}, 2023.

\bibitem{hu2021lora}
Edward~J Hu, Yelong Shen, Phillip Wallis, Zeyuan Allen-Zhu, Yuanzhi Li, Shean Wang, Lu~Wang, and Weizhu Chen.
\newblock Lora: Low-rank adaptation of large language models.
\newblock {\em arXiv preprint arXiv:2106.09685}, 2021.

\bibitem{zhang2022consecutive}
Tong Zhang, Peng Gao, Hao Dong, Yin Zhuang, Guanqun Wang, Wei Zhang, and He~Chen.
\newblock Consecutive pre-training: A knowledge transfer learning strategy with relevant unlabeled data for remote sensing domain.
\newblock {\em Remote Sensing}, 14(22):5675, 2022.

\bibitem{kirillov2023segment}
Alexander Kirillov, Eric Mintun, Nikhila Ravi, Hanzi Mao, Chloe Rolland, Laura Gustafson, Tete Xiao, Spencer Whitehead, Alexander~C Berg, Wan-Yen Lo, et~al.
\newblock Segment anything.
\newblock {\em arXiv preprint arXiv:2304.02643}, 2023.

\bibitem{radford2018improving}
Alec Radford, Karthik Narasimhan, Tim Salimans, Ilya Sutskever, et~al.
\newblock Improving language understanding by generative pre-training.
\newblock 2018.

\bibitem{radford2021learning}
Alec Radford, Jong~Wook Kim, Chris Hallacy, Aditya Ramesh, Gabriel Goh, Sandhini Agarwal, Girish Sastry, Amanda Askell, Pamela Mishkin, Jack Clark, et~al.
\newblock Learning transferable visual models from natural language supervision.
\newblock In {\em International conference on machine learning}, pages 8748--8763. PMLR, 2021.

\bibitem{ouyang2022training}
Long Ouyang, Jeffrey Wu, Xu~Jiang, Diogo Almeida, Carroll Wainwright, Pamela Mishkin, Chong Zhang, Sandhini Agarwal, Katarina Slama, Alex Ray, et~al.
\newblock Training language models to follow instructions with human feedback.
\newblock {\em Advances in Neural Information Processing Systems}, 35:27730--27744, 2022.

\bibitem{taori2023alpaca}
Rohan Taori, Ishaan Gulrajani, Tianyi Zhang, Yann Dubois, Xuechen Li, Carlos Guestrin, Percy Liang, and Tatsunori~B Hashimoto.
\newblock Alpaca: A strong, replicable instruction-following model.
\newblock {\em Stanford Center for Research on Foundation Models. https://crfm. stanford. edu/2023/03/13/alpaca. html}, 3(6):7, 2023.

\bibitem{chiang2023vicuna}
Wei-Lin Chiang, Zhuohan Li, Zi~Lin, Ying Sheng, Zhanghao Wu, Hao Zhang, Lianmin Zheng, Siyuan Zhuang, Yonghao Zhuang, Joseph~E Gonzalez, et~al.
\newblock Vicuna: An open-source chatbot impressing gpt-4 with 90\%* chatgpt quality.
\newblock {\em See https://vicuna. lmsys. org (accessed 14 April 2023)}, 2023.

\bibitem{peng2023instruction}
Baolin Peng, Chunyuan Li, Pengcheng He, Michel Galley, and Jianfeng Gao.
\newblock Instruction tuning with gpt-4.
\newblock {\em arXiv preprint arXiv:2304.03277}, 2023.

\bibitem{chen2022visualgpt}
Jun Chen, Han Guo, Kai Yi, Boyang Li, and Mohamed Elhoseiny.
\newblock Visualgpt: Data-efficient adaptation of pretrained language models for image captioning.
\newblock In {\em Proceedings of the IEEE/CVF Conference on Computer Vision and Pattern Recognition}, pages 18030--18040, 2022.

\bibitem{li2022blip}
Junnan Li, Dongxu Li, Caiming Xiong, and Steven Hoi.
\newblock Blip: Bootstrapping language-image pre-training for unified vision-language understanding and generation.
\newblock In {\em International Conference on Machine Learning}, pages 12888--12900. PMLR, 2022.

\bibitem{Google2023Bard}
Google.
\newblock Bard.
\newblock \url{https://bard.google.com/}, 2023.

\bibitem{ye2023mplug}
Qinghao Ye, Haiyang Xu, Guohai Xu, Jiabo Ye, Ming Yan, Yiyang Zhou, Junyang Wang, Anwen Hu, Pengcheng Shi, Yaya Shi, et~al.
\newblock mplug-owl: Modularization empowers large language models with multimodality.
\newblock {\em arXiv preprint arXiv:2304.14178}, 2023.

\bibitem{wei2023lenna}
Fei Wei, Xinyu Zhang, Ailing Zhang, Bo~Zhang, and Xiangxiang Chu.
\newblock Lenna: Language enhanced reasoning detection assistant, 2023.

\bibitem{lin2023sphinx}
Ziyi Lin, Chris Liu, Renrui Zhang, Peng Gao, Longtian Qiu, Han Xiao, Han Qiu, Chen Lin, Wenqi Shao, Keqin Chen, et~al.
\newblock Sphinx: The joint mixing of weights, tasks, and visual embeddings for multi-modal large language models.
\newblock {\em arXiv preprint arXiv:2311.07575}, 2023.

\bibitem{bai2023qwen}
Jinze Bai, Shuai Bai, Shusheng Yang, Shijie Wang, Sinan Tan, Peng Wang, Junyang Lin, Chang Zhou, and Jingren Zhou.
\newblock Qwen-vl: A frontier large vision-language model with versatile abilities.
\newblock {\em arXiv preprint arXiv:2308.12966}, 2023.

\bibitem{lai2023lisa}
Xin Lai, Zhuotao Tian, Yukang Chen, Yanwei Li, Yuhui Yuan, Shu Liu, and Jiaya Jia.
\newblock Lisa: Reasoning segmentation via large language model.
\newblock {\em arXiv preprint arXiv:2308.00692}, 2023.

\bibitem{hu2023rsgpt}
Yuan Hu, Jianlong Yuan, Congcong Wen, Xiaonan Lu, and Xiang Li.
\newblock Rsgpt: A remote sensing vision language model and benchmark.
\newblock {\em arXiv preprint arXiv:2307.15266}, 2023.

\bibitem{kuckreja2023geochat}
Kartik Kuckreja, Muhammad~Sohail Danish, Muzammal Naseer, Abhijit Das, Salman Khan, and Fahad~Shahbaz Khan.
\newblock Geochat: Grounded large vision-language model for remote sensing.
\newblock {\em arXiv preprint arXiv:2311.15826}, 2023.

\bibitem{girshick2014rich}
Ross Girshick, Jeff Donahue, Trevor Darrell, and Jitendra Malik.
\newblock Rich feature hierarchies for accurate object detection and semantic segmentation.
\newblock In {\em Proceedings of the IEEE conference on computer vision and pattern recognition}, pages 580--587, 2014.

\bibitem{xie2021oriented}
Xingxing Xie, Gong Cheng, Jiabao Wang, Xiwen Yao, and Junwei Han.
\newblock Oriented r-cnn for object detection.
\newblock In {\em Proceedings of the IEEE/CVF international conference on computer vision}, pages 3520--3529, 2021.

\bibitem{girshick2015fast}
Ross Girshick.
\newblock Fast r-cnn.
\newblock In {\em Proceedings of the IEEE international conference on computer vision}, pages 1440--1448, 2015.

\bibitem{ding2019learning}
Jian Ding, Nan Xue, Yang Long, Gui-Song Xia, and Qikai Lu.
\newblock Learning roi transformer for oriented object detection in aerial images.
\newblock In {\em Proceedings of the IEEE/CVF Conference on Computer Vision and Pattern Recognition}, pages 2849--2858, 2019.

\bibitem{dai2022ao2}
Linhui Dai, Hong Liu, Hao Tang, Zhiwei Wu, and Pinhao Song.
\newblock Ao2-detr: Arbitrary-oriented object detection transformer.
\newblock {\em IEEE Transactions on Circuits and Systems for Video Technology}, 2022.

\bibitem{xiao2023clip}
Linhui Xiao, Xiaoshan Yang, Fang Peng, Ming Yan, Yaowei Wang, and Changsheng Xu.
\newblock Clip-vg: Self-paced curriculum adapting of clip via exploiting pseudo-language labels for visual grounding.
\newblock {\em arXiv preprint arXiv:2305.08685}, 2023.

\bibitem{zhan2023rsvg}
Yang Zhan, Zhitong Xiong, and Yuan Yuan.
\newblock Rsvg: Exploring data and models for visual grounding on remote sensing data.
\newblock {\em IEEE Transactions on Geoscience and Remote Sensing}, 61:1--13, 2023.

\bibitem{oquab2023dinov2}
Maxime Oquab, Timoth{\'e}e Darcet, Th{\'e}o Moutakanni, Huy Vo, Marc Szafraniec, Vasil Khalidov, Pierre Fernandez, Daniel Haziza, Francisco Massa, Alaaeldin El-Nouby, et~al.
\newblock Dinov2: Learning robust visual features without supervision.
\newblock {\em arXiv preprint arXiv:2304.07193}, 2023.

\bibitem{chen2015microsoft}
Xinlei Chen, Hao Fang, Tsung-Yi Lin, Ramakrishna Vedantam, Saurabh Gupta, Piotr Doll{\'a}r, and C~Lawrence Zitnick.
\newblock Microsoft coco captions: Data collection and evaluation server.
\newblock {\em arXiv preprint arXiv:1504.00325}, 2015.

\bibitem{he2022masked}
Kaiming He, Xinlei Chen, Saining Xie, Yanghao Li, Piotr Doll{\'a}r, and Ross Girshick.
\newblock Masked autoencoders are scalable vision learners.
\newblock In {\em Proceedings of the IEEE/CVF conference on computer vision and pattern recognition}, pages 16000--16009, 2022.

\bibitem{chen2023rsprompter}
Keyan Chen, Chenyang Liu, Hao Chen, Haotian Zhang, Wenyuan Li, Zhengxia Zou, and Zhenwei Shi.
\newblock Rsprompter: Learning to prompt for remote sensing instance segmentation based on visual foundation model.
\newblock {\em arXiv preprint arXiv:2306.16269}, 2023.

\bibitem{han2021fine}
Yaqi Han, Xinyi Yang, Tian Pu, and Zhenming Peng.
\newblock Fine-grained recognition for oriented ship against complex scenes in optical remote sensing images.
\newblock {\em IEEE Transactions on Geoscience and Remote Sensing}, 60:1--18, 2021.

\bibitem{9560031}
Jian Ding, Nan Xue, Gui-Song Xia, Xiang Bai, Wen Yang, Michael Yang, Serge Belongie, Jiebo Luo, Mihai Datcu, Marcello Pelillo, and Liangpei Zhang.
\newblock Object detection in aerial images: A large-scale benchmark and challenges.
\newblock {\em IEEE Transactions on Pattern Analysis and Machine Intelligence}, pages 1--1, 2021.

\bibitem{zhang2021sar}
Tianwen Zhang, Xiaoling Zhang, Jianwei Li, Xiaowo Xu, Baoyou Wang, Xu~Zhan, Yanqin Xu, Xiao Ke, Tianjiao Zeng, Hao Su, et~al.
\newblock Sar ship detection dataset (ssdd): Official release and comprehensive data analysis.
\newblock {\em Remote Sensing}, 13(18):3690, 2021.

\bibitem{wei2020hrsid}
Shunjun Wei, Xiangfeng Zeng, Qizhe Qu, Mou Wang, Hao Su, and Jun Shi.
\newblock Hrsid: A high-resolution sar images dataset for ship detection and instance segmentation.
\newblock {\em Ieee Access}, 8:120234--120254, 2020.

\bibitem{kingma2014adam}
Diederik~P Kingma and Jimmy Ba.
\newblock Adam: A method for stochastic optimization.
\newblock {\em arXiv preprint arXiv:1412.6980}, 2014.

\bibitem{liu2023remoteclip}
Fan Liu, Delong Chen, Zhangqingyun Guan, Xiaocong Zhou, Jiale Zhu, and Jun Zhou.
\newblock Remoteclip: A vision language foundation model for remote sensing.
\newblock {\em arXiv preprint arXiv:2306.11029}, 2023.

\bibitem{liu2023grounding}
Shilong Liu, Zhaoyang Zeng, Tianhe Ren, Feng Li, Hao Zhang, Jie Yang, Chunyuan Li, Jianwei Yang, Hang Su, Jun Zhu, et~al.
\newblock Grounding dino: Marrying dino with grounded pre-training for open-set object detection.
\newblock {\em arXiv preprint arXiv:2303.05499}, 2023.

\bibitem{zhao2024open}
Xiangyu Zhao, Yicheng Chen, Shilin Xu, Xiangtai Li, Xinjiang Wang, Yining Li, and Haian Huang.
\newblock An open and comprehensive pipeline for unified object grounding and detection.
\newblock {\em arXiv preprint arXiv:2401.02361}, 2024.

\bibitem{han2021align}
Jiaming Han, Jian Ding, Jie Li, and Gui-Song Xia.
\newblock Align deep features for oriented object detection.
\newblock {\em IEEE Transactions on Geoscience and Remote Sensing}, 60:1--11, 2021.

\bibitem{guo2021beyond}
Zonghao Guo, Chang Liu, Xiaosong Zhang, Jianbin Jiao, Xiangyang Ji, and Qixiang Ye.
\newblock Beyond bounding-box: Convex-hull feature adaptation for oriented and densely packed object detection.
\newblock In {\em Proceedings of the IEEE/CVF conference on Computer Vision and Pattern Recognition}, pages 8792--8801, 2021.

\bibitem{li2022oriented}
Wentong Li, Yijie Chen, Kaixuan Hu, and Jianke Zhu.
\newblock Oriented reppoints for aerial object detection.
\newblock In {\em Proceedings of the IEEE/CVF conference on computer vision and pattern recognition}, pages 1829--1838, 2022.

\bibitem{hou2022shape}
Liping Hou, Ke~Lu, Jian Xue, and Yuqiu Li.
\newblock Shape-adaptive selection and measurement for oriented object detection.
\newblock In {\em Proceedings of the AAAI Conference on Artificial Intelligence}, 2022.

\bibitem{yang2021r3det}
Xue Yang, Junchi Yan, Ziming Feng, and Tao He.
\newblock R3det: Refined single-stage detector with feature refinement for rotating object.
\newblock In {\em Proceedings of the AAAI Conference on Artificial Intelligence}, volume~35, pages 3163--3171, 2021.

\end{thebibliography}

\end{document}